\documentclass[letterpaper, 10 pt, conference]{ieeeconf}  
\IEEEoverridecommandlockouts                              
\usepackage{graphicx}
\usepackage{amsmath}
\usepackage{booktabs}   
\usepackage{multirow}   
\usepackage{array}      
\usepackage{caption}
\usepackage{float}
\usepackage{colortbl}
\usepackage{bm}
\usepackage{amssymb}
\usepackage{xcolor}
\usepackage{bbding}  
\usepackage{pifont}  
\usepackage{subfig}
\usepackage[colorlinks = true]{hyperref}
\usepackage{cite}

\newcommand{\etal}{\textit{et al}. }

\newcommand{\eg}{\textit{e}.\textit{g}. }

\definecolor{tablecolor}{HTML}{6cfe00} 
\definecolor{tablecolor2}{HTML}{d1ffaf}
\definecolor{citecolor}{HTML}{fe7b5b}
\definecolor{grey}{rgb}{0.9, 0.9, 0.9}
\definecolor{gred}{rgb}{0.859,0.267,0.216}
\definecolor{ggreen}{rgb}{0.059,0.616,0.345}
\definecolor{deepblue}{HTML}{27a2c3}
\definecolor{deepred}{HTML}{fe7b5b}

\newcommand{\dd}[2]{$#1\scriptstyle{\pm#2}$}
\newcommand{\ddgf}[2]{\cellcolor{tablecolor}$\mathbf{#1\scriptstyle{\pm#2}}$}
\newcommand{\ddsc}[2]{\cellcolor{tablecolor2}$#1\scriptstyle{\pm#2}$}

\newcommand{\ccgf}[1]{\cellcolor{tablecolor}$\mathbf{#1}$}
\newcommand{\scgf}[1]{\cellcolor{tablecolor2}$#1$}
\newcommand\blfootnote[1]{%
  \begingroup
  \renewcommand\thefootnote{}\footnote{#1}%
  \addtocounter{footnote}{-1}%
  \endgroup
}

\title{\LARGE \bf
FBI: Learning Dexterous In-hand Manipulation with Dynamic Visuotactile Shortcut Policy
}

\author{Yijin Chen$^{1*}$, Wenqiang Xu$^{1*}$, Zhenjun Yu$^{1}$, Tutian Tang$^{1}$, Yutong Li$^{1}$, Siqiong Yao$^{1}$  and Cewu Lu$^{1}$
\thanks{$^{1}$Shanghai Jiao Tong University. 
$^*$ indicates equal contribution.
{\tt\small\{st.czzz, vinjohn, jeffson-yu, tttang, davidliyutong, yaosiqiong, lucewu\}@sjtu.edu.cn}}%
}%

\begin{document}

\twocolumn[{%
\renewcommand\twocolumn[1][]{#1}%
\maketitle
\begin{center}
    \vspace{-0.1in}
    \centering
    \captionsetup{type=figure}
    \includegraphics[width=\linewidth]{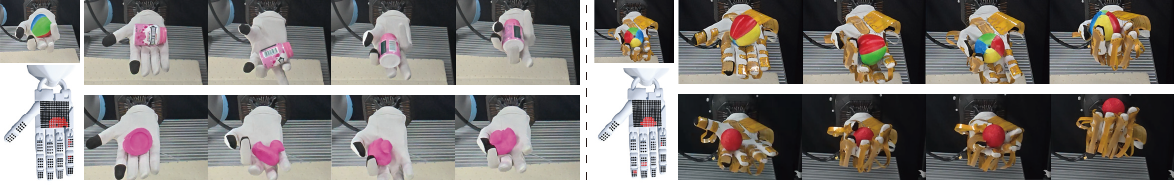}
    \captionof{figure}{We propose \textbf{Flow Before Imitation (FBI)}, a novel dynamic visuotactile imitation learning algorithm for dexterous in-hand manipulation. FBI's design enables two operational modes: with or without physical tactile sensors in the real world, largely extending the application scenarios. }
    \label{fig:teaser}
    \vspace{-0.1in}
\end{center}
}]

\blfootnote{
$^{1}$Shanghai Jiao Tong University. $^*$ indicates equal contribution. \\
{\tt\small\{st.czzz, vinjohn, jeffson-yu, tttang, davidliyutong, yaosiqiong, lucewu\}@sjtu.edu.cn}

}

\thispagestyle{empty}
\pagestyle{empty}


\begin{abstract}

Dexterous in-hand manipulation is a long-standing challenge in robotics due to complex contact dynamics and partial observability. While humans synergize vision and touch for such tasks, robotic approaches often prioritize one modality, therefore limiting adaptability. This paper introduces Flow Before Imitation (FBI), a visuotactile imitation learning framework that dynamically fuses tactile interactions with visual observations through motion dynamics. Unlike prior static fusion methods, FBI establishes a causal link between tactile signals and object motion via a dynamics-aware latent model. FBI employs a transformer-based interaction module to fuse flow-derived tactile features with visual inputs, training a one-step diffusion policy for real-time execution. Extensive experiments demonstrate that the proposed method outperforms the baseline methods in both simulation and the real world on two customized in-hand manipulation tasks and three standard dexterous manipulation tasks.
Code, models, and more results are available in the website \url{https://sites.google.com/view/dex-fbi}. 

\end{abstract}

\section{INTRODUCTION}

In-hand manipulation, which aims to reposition objects within a single dexterous hand, remains a critical unsolved problem in robot learning due to complex contact dynamics and partial observability.
Humans can easily interact with objects in hand thanks to the fusion of vision and touch. Vision tracks global object states for task completion, while tactile sensing enables precise force adjustments during contact. Existing robotic approaches often prioritize either vision \cite{xu2023dexterous,qin2022dexmv,andrychowicz2020learning} or tactile sensing \cite{yin2023rotating,yang2024anyrotate,lee2024dextouch}, limiting their effectiveness in complex manipulation scenarios. While recent visuotactile approaches \cite{qi2023general,yang2024anyrotate,guzey2023dexterity,guzey2024see,yuan2024robot} demonstrate improved manipulation stability, most rely on optical tactile sensors \cite{qi2023general,yang2024anyrotate,guzey2023dexterity,guzey2024see} whose bulkiness (typically $\geq 8mm$ thickness) prevents deployment across full-hand articulations (\eg metacarpal joints). Distributed tactile arrays (\eg piezoresistive arrays) present a practical alternative \cite{yin2023rotating,sundaram2019learning,jiang2024capturing}, offering hardware compatibility with commercial dexterous manipulators, full-palm coverage, and sufficient contact force resolution for detecting incipient slip. Despite these advantages, the fusion of distributed tactile data with visual streams remains understudied in policy learning frameworks.

Existing visuotactile fusion methods typically process multimodal inputs through \textit{static fusion}, such as introducing encoded features \cite{guzey2024see} or augmenting visual point clouds with tactile contact coordinates \cite{yuan2024robot} from a single frame. These approaches neglect the intrinsic causal relationship between tactile interactions and object state transitions during dynamic manipulation: Tactile forces drive object state changes. Therefore, we take the path of \textit{dynamic fusion}, extracting tactile information from temporal object motion flow via a dynamics-aware latent model. This representation enables two operational modes in practice: (1) Vision-Only mode. We can infer tactile cues without physical sensors; (2) Visuotactile mode. With tactile sensors, measured contact forces can be used to refine flow predictions to improve precision. Therefore, our method can be flexibly deployed to a wider range of application scenarios, even when the tactile sensor is unavailable (Figure \ref{fig:teaser}).

Building on this dynamic perspective, we present Flow Before Imitation (FBI) — an imitation learning framework integrating tactile and visual cues through motion dynamics. FBI processes multimodal inputs, including consecutive robot states, partially observed point clouds, and tactile readings, through dedicated encoder networks to extract proprioception and contact features. These encoded features are then dynamically fused and conditioned into a one-step shortcut model \cite{frans2024one} to predict action sequences. This design enables efficient, real-time policy execution while integrating tactile-visual dynamics for robust in-hand manipulation.

To evaluate our method, we test the FBI on five tasks, including in-hand reorientation, in-hand pushing, and three dexterous tasks from a public benchmark, \textbf{Adroit}~\cite{rajeswaran2017learning}, with objects ranging from the standard simple cubes to real-world items. Compared with four baseline methods including DP3 \cite{ze20243d}, ManiCM \cite{lu2024manicm}, Ada-Flow \cite{hu2024adaflow}, and Consistency Policy \cite{prasad2024consistency}, the proposed method performs better. In simulations, it reaches 64.7\%  (Vision-Only) to 66.5\%  (Visuotactile)  average success, 16.6\% to 18.4\% higher than the previous SOTA method, respectively. It is especially good at hard reorientation tasks, 19.3\% (Vision-Only) to 21.4\% (Visuotactile) higher.
We also conduct real-world tests that show similar gains, 33.5\% (Vision-Only) to 35.0\% (Visuotactile) vs. 18.5\% baseline.

Our contribution can be summarized as follows:

    1) We propose Flow Before Imitation (FBI), a visuotactile fusion method that dynamically infers tactile interactions from object motion flow, enabling real-time control via a one-step diffusion policy. It enables two operational modes with or without physical tactile sensors in the real-world experiments, largely extending the application scenarios.

    2) Extensive evaluation across five tasks in both simulation and real-world experimental settings demonstrates the FBI’s superiority over baseline methods.

\section{Related Works}

\subsection{Multimodal Sensing for Dexterous Manipulation.}
Considering that humans operate with dexterous manipulation, multimodal sensing, especially vision and touch, is of concern.
To empower robots with such abilities, researchers often start with simplified, single-modal perception settings including vision-only \cite{xu2023dexterous,qin2022dexmv,andrychowicz2020learning,arunachalam2023dexterous,handa2023dextreme} and touch-only \cite{yin2023rotating,yang2024anyrotate,lee2024dextouch}. Though much progress has been made, vision-only methods, by nature, struggle with occlusions, while tactile-only systems lack global spatial awareness.
Recent advances highlight the benefits of multimodal sensing. Guzey \etal \cite{guzey2023dexterity,guzey2024see} combined self-supervised tactile pre-training with visual inputs to address contact reasoning, outperforming single-modal baselines. Yuan \etal \cite{yuan2024robot} introduced a point cloud-based tactile representation fused with vision to enhance spatial planning. While these methods demonstrate the potential of multimodal fusion, some works exclusively use finger contacts as effectors while neglecting palm interactions~\cite{qi2023general, yang2024anyrotate}, and others restrict object rotation to predefined axes~\cite{yin2023rotating, yuan2024robot}. In contrast, our approach utilizes the full palm side to manipulate objects, enabling the ability to reorient irregular objects and execute controlled pushes. 

\subsection{Imitation Learning for Dexterous Manipulation.}
Since analytical planning \cite{cruciani2018dexterous} or control \cite{khadivar2023adaptive} has limited generalization ability to object variance, learning for dexterous manipulation gains increasing attention. While reinforcement learning (RL) \cite{yang2024anyrotate,andrychowicz2020learning,qi2023general} excels without demonstrations, it requires intricate reward design. 
On the contrary, imitation learning methods can mitigate the reward design problem and are known to have better sample efficiency with demonstration data. The policy learning methods in imitation are developed along with the generative methods, from nearest neighbors-based approaches~\cite{guzey2023dexterity,arunachalam2023dexterous,guzey2024see}, Gaussian-based dynamic motion primitives \cite{hammoud2022hand}, generative adversarial networks \cite{antotsiou2018task}, to diffusion models \cite{ze20243d, lu2024manicm,hu2024adaflow,prasad2024consistency}.
Ze \etal~\cite{ze20243d} replaced 2D image inputs in vanilla diffusion policy \cite{chi2023diffusion} with 3D point clouds, which significantly enhanced the performance of dexterous manipulation. Later, some works~\cite{hu2024adaflow, prasad2024consistency, su2024motion} have replaced the diffusion process with flow matching~\cite{lipman2022flow, frans2024one} to reduce inference time.
However, these methods only exploit a single modality.

\section{Method}
\label{sec:method}

Dexterous in-hand manipulation requires rich hand-object interactions, where vision and touch complement each other. We propose Flow Before Imitation (FBI), a visuotactile imitation learning algorithm for this task. 
FBI takes two consecutive robot's state $s_{t-1}$, $s_t\in \mathbb{R}^{N_s}$, partially-observed point cloud frames $P_{t-1}$, $P_t\in\mathbb{R}^{N_p \times 3}$ and tactile readings $R_t \in \mathbb{R}^{N_r}$ as input, where $N_s$ is the number of joints, $N_p$ is the number of points in the downsampled point clouds  (Section \ref{sec:encoders}), $N_r$ is the number of contact keypoints (Section \ref{sec:f2t}).  Such multimodal data are first processed by the multimodal encoders (Section \ref{sec:encoders}).  After that, the encoded features are fused and passed as conditions to a shortcut model \cite{frans2024one} to predict the action series $\mathbf{A}_t \in \mathbb{R}^{H \times d_a}$ (Section \ref{sec:policy}), where $H$ is the horizon of the action series, $d_a$ is the dimension of each action. 

\begin{figure*}[!tbp]
    \centering
    \includegraphics[width=0.85\linewidth]{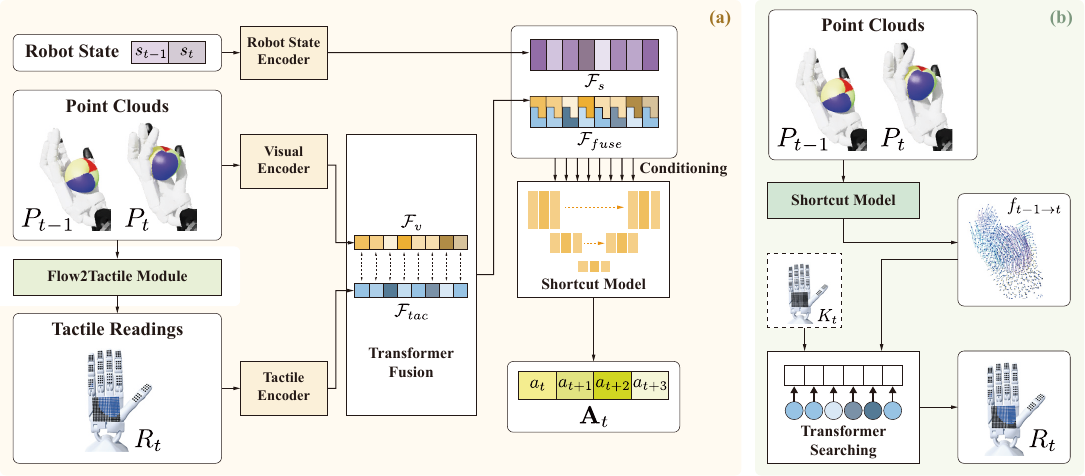}
    \caption{\textbf{Overview of FBI pipeline.} (a) The overview of our pipeline, including multimodal encoders, visuotactile feature fusing module, and the policy generation module. (b) The Flow2Tactile module that predicts the tactile readings, making it compatible with vision-only scenarios. }
    \label{fig:pipeline}
    \vspace{-0.5cm}
\end{figure*}

While visuotactile systems excel in ideal conditions, we often witness hardware limitations in real-world deployments, since distributed tactile sensors could be unavailable or cost-prohibitive. To enhance accessibility and lower the barriers for labs and industries without tactile hardware, our method can maintain the dexterous manipulation capabilities using only visual input. Thanks to the dynamic perspective, tactile readings $R_t$ can be predicted from the point cloud flow $f_{t-1\rightarrow t}\in \mathbb{R}^{N_p \times 3}$ by the Flow2Tactile Module (Section \ref{sec:f2t}). An overview of our method is shown in Figure~\ref{fig:pipeline}.

\subsection{Multimodal Encoders}\label{sec:encoders} 
\subsubsection{Robot State Encoder} We use a 3-layer MLP as the state encoder $\bm{f}_s(\cdot)$ to encode the robot's two consecutive proprioception $s_{t-1}$, $s_t$ into a compact state feature $\mathcal{F}_s \in \mathbb{R}^{d_s}$.

\subsubsection{Visual Encoder} Given the two-frame point clouds of the hand and the object, we first crop and downsample them into $N_p$ points and forward them to a lightweight MLP encoder $\bm{f}_v(\cdot)$. The lightweight encoder, consisting of a three-layer MLP, a max-pooling function, and an output MLP layer for feature dimension reduction, takes the downsampled point cloud $P_{t-1}$ and $P_t$ as input and outputs the compact feature $\mathcal{F}_v \in \mathbb{R}^{d_v}$, where $d_v$ is the dimension of the visual feature. The visual feature will be fused with the tactile feature to form the visuotactile feature.

\subsubsection{Tactile Encoder} To process the tactile readings, we use a four-layer MLP as the tactile encoder $\bm{f}_t(\cdot)$ to encode tactile feature $\mathcal{F}_{tac} \in \mathbb{R}^{d_{tac}}$, where $d_{tac}$ is the dimension of the tactile feature. Note that in the vision-only setting, the tactile readings come from the prediction results of the Flow2Tactile module. In the visuotactile setting, where the physical tactile sensor is available, it comes as direct readings from the sensors.

\subsection{Visuotactile Policy Generation} \label{sec:policy}
With the visual feature $\mathcal{F}_v$ and tactile feature $\mathcal{F}_{tac}$, we are able to obtain the visuotactile features and use them as conditions for the policy generation model. In this work, we adopt a shortcut model~\cite{frans2024one} to predict action series $\mathbf{A}_t$. The shortcut model is a flow-matching-based generative model that enables high-quality one-step sampling. In the training phase, starting from an initial inference step $dt$, it optimizes both flow matching loss and self-consistency loss simultaneously to improve the model's performance and increase the effective inference step size at the same time, ultimately achieving single-step generation ($dt = 1$). In the sampling phase, given condition $C$ and a desired inference step size $dt$, the traditional flow-matching model can generate target distributions $p_{\theta}(x | C)$ from a standard normal distribution $x \sim \mathcal{N}(0, I)$. Starting from a Gaussian noise $x^0$, it utilizes a U-net~\cite{ronneberger2015u} as the velocity predicting network $v_{\theta}(\cdot)$ to recursively predicts the velocity at time step $t \in [0, 1)$ until finally reach the clear action $x^1$ at denoising time step 1:
\begin{equation} \label{eq:fm_inference}
    x^{t+dt} = x^t + v_{\theta}(x^t, t, dt, C)dt.
\end{equation}
In the shortcut model, which is one-step generation, we have $t=0, dt=1$. 

\subsubsection{Fusing Tactile Features with Vision Features}   The encoded visual feature $\mathcal{F}_v$ and tactile feature $\mathcal{F}_{tac}$  are forwarded to a transformer fusion module $T_f(\cdot)$ which will take $\mathcal{F}_v$ as the template feature, $\mathcal{F}_{tac}$ as the searching feature and outputs the visuotactile feature. Considering visual observation is also crucial for environmental perception, we concatenate the visuotactile feature with the visual feature to form the final fused feature $\mathcal{F}_{fuse}$. 

\subsubsection{Predicting Actions from Multimodal Conditions}
In our case, the model condition $C$ consists of $\mathcal{F}_{fuse}$ and the robot state features $\mathcal{F}_s$. The velocity output by $v_\theta(\cdot)$ represents the denoising direction from the noisy action series $\mathbf{A}^0$ to the predicted action series $\mathbf{A}_t$. With the implementation of the shortcut model, we can use the $v_\theta(\cdot)$ \textbf{only once} to complete the prediction:
\begin{equation}
    \mathbf{A}_t = \mathbf{A}^t + v_{\theta}(\mathbf{A}^t, t, dt,  \mathcal{F}_s, \mathcal{F}_{fuse})dt,
\end{equation}

where $t = 0, dt = 1$ in our case. The $t$ in $\mathbf{A}_t$ denotes the time frame in the task progress, while the $t$ in $\mathbf{A}^t$ denotes the denoising time step. The benefits of the feature fusion on our policy will be included in Section~\ref{sec:experiment results}.

\subsubsection{Training Objective} 
We follow the main idea of training shortcut models to train our shortcut policy for one-step action generation. We first determine a minimum inference step size $dt$ and optimize both \textbf{flow matching loss} and \textbf{self-consistency loss} simultaneously. Assume that $t \in [0, 1)$ is the randomly selected variable, $\mathbf{A}^0 \sim \mathcal{N}(0, I)$ is the starting action Gaussian noise, $\mathbf{A}^1$ is the ground-truth action series, $\mathcal{F}=\{\mathcal{F}_s, \mathcal{F}_{fuse}\}$ is the conditional feature during the training phase, $v_{\theta}(\cdot)$ is the velocity predicting model, then the loss function should be given as:
\begin{equation}
\begin{aligned}
\label{eq:loss_policy}
    \mathcal{L}_{FM} & = MSE\left( (\mathbf{A}^1 - \mathbf{A}^0), v_{\theta}(\mathbf{A}^t, t, dt, \mathcal{F}) \right) \\
    \mathcal{L}_{SC} & = MSE\left( v_{\theta}(\mathbf{A}^t, t, 2dt,\mathcal{F}), v_{target} \right) \\
    \mathcal{L} & = \mathcal{L}_{FM} + \mathcal{L}_{SC} ,
\end{aligned}
\end{equation}
where $v_{target}= \left[ v_{\theta}(\mathbf{A}^t, t, dt,\mathcal{F}) + v_{\theta}(\mathbf{A}^{t+dt},t+dt,dt,\mathcal{F})\right] / {2} $ and $\mathbf{A}^{t+dt}=\mathbf{A}^{t} +v_{\theta}(\mathbf{A}^t,t,dt,\mathcal{F})dt$.

\subsection{Flow2Tactile Module} \label{sec:f2t}

Dense contact information is crucial for contact-rich manipulation tasks, yet acquiring it demands high-precision tactile sensors, which are often unavailable in real-world experiments. Therefore, we leverage a \textbf{Flow2Tactile module} to predict dense contact states using object state flows. In doing so, the framework can also work in the vision-only mode.

\begin{figure}[!htbp]
    \centering
    \vspace{-0.28cm}
    \includegraphics[width=1.0\linewidth]{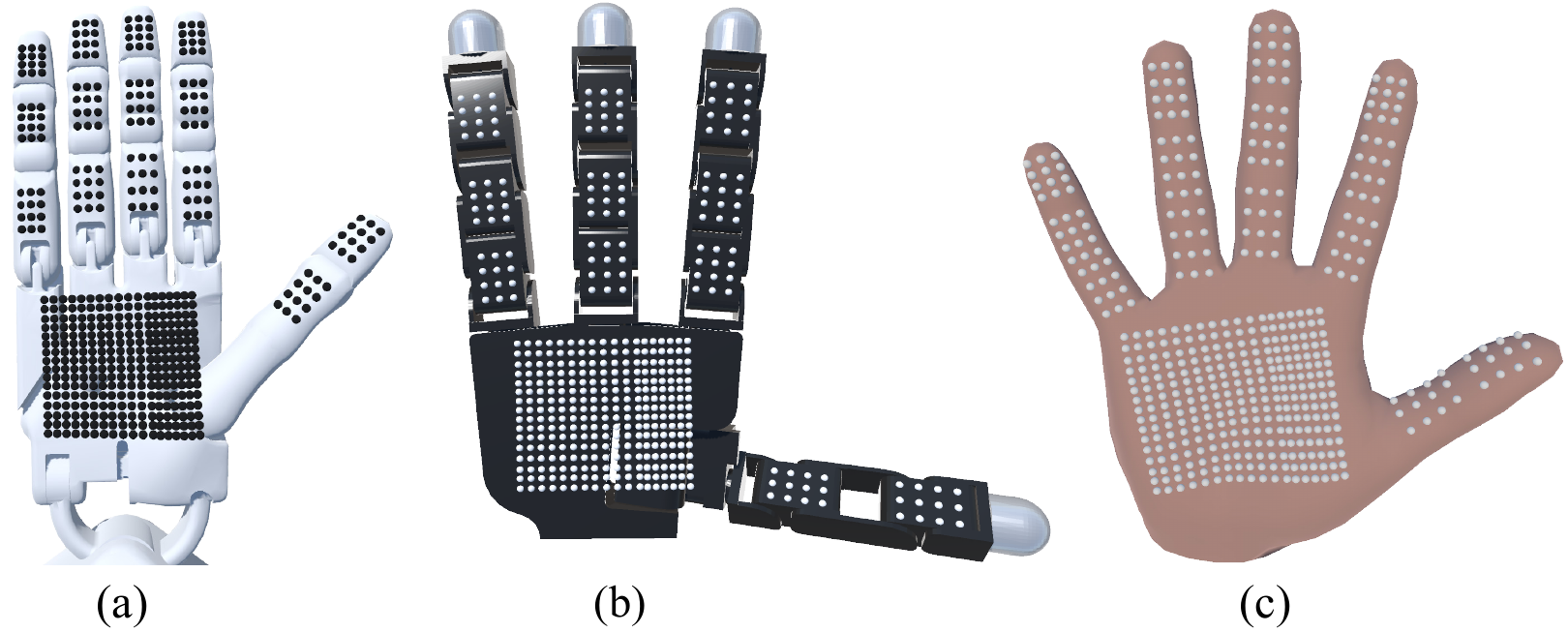}
    \caption{Contact keypoint designs for (a): Shadow Hand~\cite{shadowhand}, (b): Allegro Hand~\cite{allegro}, (c): Mano Hand~\cite{sundaram2019learning}. }
    \label{fig:keypoint_design.pdf}
    \vspace{-0.4cm}
\end{figure}

Specifically, an object state flow $f_{t-1\rightarrow t}$ is first predicted by a shortcut model using two frames point clouds $P_{t-1}$ and $P_t$ to represent how $P_{t-1}$ transforms into $P_t$. Then, a pre-trained transformer searching model $T_s(\cdot)$ predicts the tactile readings $R_t$ on a pre-defined layout of contact keypoints utilizing $f_{t-1\rightarrow t}$ and the current coordinate of contact keypoints $K_t\in\mathbb{R}^{N_k\times3}$, obtaining the dense contact state at time step $t$. Here, $N_k$ is the number of contact keypoints. The formation of the tactile processing module should be given as:
\begin{equation}
    R_t = T_s(K_t, f_{t-1\rightarrow t}).
\end{equation}
The design of these contact keypoints aims to comprehensively cover the hand's palm side while referring to the physical parameters of sensors in the real world, so that real sensors can fit in the layout by finding a suitable correspondence between the real tactile sensor layout and the predefined contact keypoint layout. Take the Shadow Hand~\cite{shadowhand} as an example, a set of 456 points, consisting of 12 keypoints on each finger link and 288 keypoints on the palm, is used as our contact keypoints (Figure ). We use forward kinematics to calculate the current coordinates $K_t\in \mathbb{R}^{456\times3}$ of the contact keypoints at time step $t$. We refer to some distributed tactile sensor designs~\cite{liu2017glove, 
 yin2023rotating,sundaram2019learning, jiang2024capturing} and design keypoint layout for several hands. We display them in Figure~\ref{fig:keypoint_design.pdf}. 
 
\subsubsection{Training Objectives}
The pre-trained Flow2Tactile module contains two parts: the flow generation part and the transformer searching module. We give the training losses directly:

\begin{equation}
\begin{aligned}
    \mathcal{L}_{FM} &= CD\left(P_t, (P_{t-1} + f_{t-1\rightarrow t})\right) \\
    \mathcal{L}_{SC} &= MSE\left( v_{\theta}(f^t, t, 2dt,P), v_{target} \right) \\
    \mathcal{L} &= \mathcal{L}_{FM} + \mathcal{L}_{SC}, 
\end{aligned}
\end{equation}

where $CD$ stands for Chamfer Distance, $v_{target}$ is alike Eq.\ref{eq:loss_policy}, only to substitute $\mathbf{A}^t$ to $f^t$, $\mathcal{F}$ to $P$.

For the transformer searching module, we compute the loss between the output and the ground-truth tactile readings $R_{gt}$, which is calculated in simulation (detailed in Section~\ref{sec:training_and_evaluation_in_simulation}). The loss function is given as:
\begin{equation}
    \mathcal{L}_t = MSE\left( R_{gt},  T_s(K_t, f_{t-1 \rightarrow t}) \right).
\end{equation}

\section{Experiments}
\label{Experiments}
In this section, we systematically evaluate our method in both simulation and real experiments. Specifically, we evaluate the FBI in 2 in-hand-manipulation tasks on various objects in both simulation and real-world settings. We also select three dexterous manipulation tasks from a public benchmark for fair comparison with baseline methods. 

\begin{table*}[!htbp]
\centering
\vspace{-0.02in}

\resizebox{1.0\textwidth}{!}{%
\begin{tabular}{l|c|ccccccccc|c|ccc|cc}
\toprule

    & \multicolumn{1}{c|}{}& \multicolumn{9}{c|}{In-hand Reorientation} & \multicolumn{1}{c|}{In-hand Push} & \multicolumn{3}{c|}{Adroit} & \\
  Algorithm $\backslash$ Task &  NFE &Cube & Apple & Vase  & Ring  & Duck & Owl & daily object1 & daily object2 & daily object3 & Balls & Hammer & Door & Pen & \textbf{Average} \\

\midrule

\textbf{FBI (Vision-Only)} & 1 & \ddsc{80}{3} & \ddsc{86}{2} & \ddsc{74}{5} & \ddgf{76}{1}  & \ddgf{28}{5} & \ddsc{27}{4} & \ddsc{65}{4} & \ddsc{39}{2} & \ddgf{25}{7} & \ddsc{95}{3} & \ddgf{100}{0} & \ddgf{77}{3} & \ddsc{69}{4} & \scgf{64.7} \\

\textbf{FBI (Visuotactile)} & 1 & \ddgf{84}{4} & \ddgf{90}{2} & \ddgf{80}{2} & \ddgf{76}{8}  & \ddgf{28}{2} &  \ddgf{29}{3} & \ddgf{68}{2} & \ddgf{42}{5} & \ddsc{22}{10} & \ddgf{97}{1} & \ddgf{100}{0} & \ddsc{75}{2} & \ddgf{73}{3} & \ccgf{66.5}  \\ 

DP3 &  10  & \dd{51}{9} & \dd{53}{12} & \dd{44}{2}	& \dd{50}{5} & \dd{20}{5} & \dd{18}{3} &   \dd{49}{2} &  \dd{24}{5} &  \dd{18}{6} &\dd{84}{7}  & \ddgf{100}{0} & \dd{69}{5} & \dd{45}{5} & $48.1$ \\

ManiCM & 1  & \dd{50}{3} & \dd{50}{10} & \dd{46}{3} & \dd{53}{2} & \dd{17}{2} &  \dd{15}{6} &  \dd{47}{6} &  \dd{25}{7} &  \dd{16}{4} & \dd{86}{5} & \ddgf{100}{0} & \dd{68}{3} & \dd{43}{6} & $47.4$  \\
Ada-Flow & 1.82 & \dd{15}{12} & \dd{30}{2} & \dd{21}{4} & \dd{15}{7} & \dd{1}{1} &  \dd{3}{2} & \dd{30}{7} & \dd{25}{2} &  \dd{10}{3} & \dd{44}{15} & \dd{50}{12} & \dd{52}{3} & \dd{28}{4} & $24.9$ \\
Consistency Policy &  1 & \dd{19}{8} & \dd{32}{3} & \dd{20}{3} & \dd{18}{6} & \dd{1}{2} & \dd{2}{0} & \dd{25}{10} & \dd{26}{3} &  \dd{12}{3} & \dd{48}{10} & \dd{55}{6} & \dd{50}{7} & \dd{26}{5} & $25.7$ \\

\bottomrule
\end{tabular}}
\caption{\textbf{Simulation Results.} All baselines are run under the same simulation parameters for fair comparison. Lighter green indicates the second-best.}
\label{table: simulation results}
\vspace{-0.02in}
\end{table*}

\begin{table*}[!htbp]
\centering
\vspace{-0.02in}
\resizebox{1.0\textwidth}{!}{%
\begin{tabular}{l|c|ccccccccc|c|cc}
\toprule

    & \multicolumn{1}{c|}{}& \multicolumn{9}{c|}{In-hand Reorientation} & \multicolumn{1}{c|}{In-hand Push} & \\
  Algorithm $\backslash$ Task &  NFE &Cube & Apple & Vase  & Ring & Duck & Owl & daily object1 & daily object2 & daily object3 & Balls & \textbf{Average} \\

\midrule

\textbf{FBI (Vision-Only)} & 1 & \ccgf{45.0} & \ccgf{50.0} & \scgf{35.0}  & \ccgf{40.0}  & \ccgf{20.0} & \scgf{15.0} & \ccgf{30.0} & \ccgf{25.0} & \scgf{15.0} & \scgf{60.0} & \scgf{33.5} \\

\textbf{FBI (Visuotactile)} & 1 & \ccgf{45.0} & \ccgf{50.0} & \ccgf{40.0}  & \ccgf{40.0}  &\scgf{15.0} & \ccgf{20.0} & \ccgf{30.0} & \ccgf{25.0} & \ccgf{20.0} & \ccgf{65.0} & \ccgf{35.0} \\

DP3 &  10  & $10.0$ & $15.0$ & $5.0$ & $5.0$ &	$0.0$ & $0.0$ & $10.0$  & $5.0$ & $0.0$ & $20.0$ & $7.0$\\

ManiCM & 1  & $25.0$ & 25.0 & $30.0$ & $15.0$ & $10.0$ &  $5.0$ & $20.0$ & $10.0$ & \scgf{15.0} & $30.0$ & $18.5$  \\
Ada-Flow & 1.32 & $10.0$ & $10.0$ & $10.0$ & $15.0$   & $0.0$ &  $0.0$  & $15.0$ & $15.0$ &  $5.0$ & $25.0$ & $10.5$ \\
Consistency Policy &  1 & $15.0$ & $10.0$ & $15.0$ & $15.0$ &  $5.0$ & $0.0$ & $10.0$ & $15.0$ & $10.0$ & $20.0$ & $11.5$ \\

\bottomrule
\end{tabular}}
\caption{\textbf{Real-World Experiment Results.} FBI outperforms all baselines in all tasks, achieving 15.0\% (Vision-Only) to 16.5\% (Visuotactile) improvement compared to the previous SOTA baseline, ManiCM. Lighter green indicates the second-best.}
\label{table:real world experiments}
\vspace{-0.02in}
\end{table*}

\subsection{Task Setup} \label{sec:task description}

\begin{figure}[!htbp]
    \centering
    \vspace{-0.2cm}
    \includegraphics[width=1.0\linewidth]{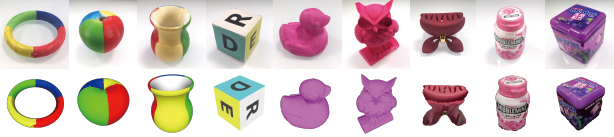}
    \caption{Object dataset appearance in the real world (top row) and simulation (bottom row).}
    \label{fig:object_dataset.pdf}
    \vspace{-0.5cm}
\end{figure}

\subsubsection{Object Dataset} An overview of our object dataset is shown in Figure~\ref{fig:object_dataset.pdf}. The object dataset contains objects of various shapes, weights, and colors. It has 6 3D-printed objects of both simple and complex shapes. The symmetric shapes are colored to indicate the orientation. We also have 3 daily objects to test the adaptability of our algorithm. 

\subsubsection{Robot Hand} For both simulation and real-world experiments, we utilized the Shadow Hand~\cite{shadowhand}, a five-fingered robotic hand with 24 degrees of freedom.

\subsubsection{Task Description}We evaluate FBI on two in-hand tasks: \textbf{Push} (moving objects to target positions) and \textbf{Reorientation} (adjusting orientations to target pose) in both simulation and real-world. Three Adroit tasks (\textbf{Door}, \textbf{Hammer}, \textbf{Pen})~\cite{rajeswaran2017learning} using Shadow Hand in MuJoCo~\cite{todorov2012mujoco} are also selected, following prior benchmarks~\cite{ze20243d,lu2024manicm}.

\subsubsection{Evaluation Metric and Goal Definition} Success Rate (SR) is used for all tasks. Push success: object-target position $\leq 1$mm; Reorientation success: orientation difference $\leq 0.1$rad; Adroit tasks retain original criteria. SR is auto-calculated in simulation and manually assessed in the real world. In our own tasks, the target goal is a synthetic object point cloud transformed into the desired pose, which is concatenated with the observed scene point cloud and input to the model. For Adroit tasks, goals (door, hammer) are involved in the scene point cloud. 

\subsection{Training and Evaluation in Simulation}\label{sec:training_and_evaluation_in_simulation}
We simulate \textbf{In-hand Push} and \textbf{In-hand Reorientation} tasks using Isaac Sim 4.1.0~\cite{nvidia2022isaacsim} (25 Hz control, 120 Hz simulation). State-based PPO agents~\cite{schulman2017proximal}, trained with learning rate 5e-4 and training epochs 1e4, achieving $65.7\%$ success rate for in-hand reorientation and $96.8\%$ for in-hand push, are used to collect 5000 success trajectories per object. For \textbf{Adroit} tasks, in MuJoCo, a VRL3~\cite{wang2022vrl3} agent generates 10 success trajectories per task. FBI is trained for 300 epochs on the first two tasks and 3000 epochs on Adroit to ensure convergence. 

\begin{figure}[!htbp]
    \centering
    \vspace{-0.28cm}
    \includegraphics[width=1.0\linewidth]{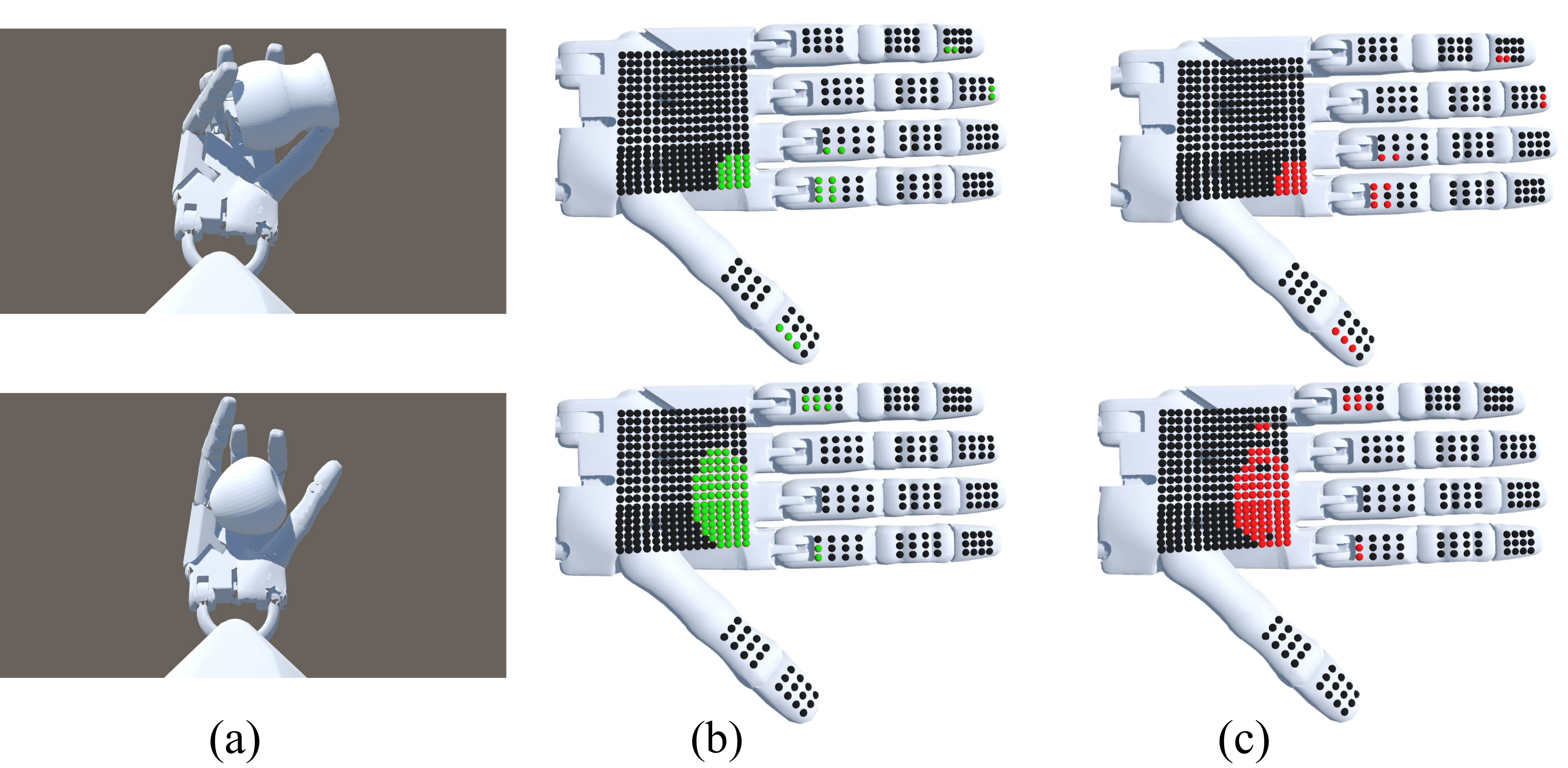}
    \caption{Qualitative results of the Flow2Tactile Module. (a): The rendering result. (b): Ground-truth tactile readings (ZeMa). (c): Predicted tactile readings (Flow2Tactile Module).}
    \label{fig:zema_vis.pdf}
    \vspace{-0.4cm}
\end{figure}

To train the Flow2Tactile Module, we use ZeMa~\cite{du2024intersection} to calculate the ground-truth tactile readings in simulation, because Isaac Sim doesn't currently support adding dense tactile sensors to robot hands and querying tactile readings. ZeMa~\cite{du2024intersection} is a simulation platform based on the finite element method (FEM), featuring the most accurate contact modeling presently. It has been employed in some studies~\cite{jiang2024capturing, ma2025grip, yu2024dynamic} to replicate real-world contact scenarios. Thus, we can treat the contact states calculated by ZeMa as ground-truth tactile readings. Specifically, we first collect the correlated states of hands and objects in the RL-generated demonstrations. Then, we replay the trajectories in ZeMa, calculate the frame-wise tactile readings using incremental potential contact (IPC), and finally binarize them to obtain the ground-truth tactile readings. We choose $1000$ trajectories per object and the average object density is $463.1 kg/m^3$. After training, our Flow2Tactile can reach an $85.5\%$ prediction accuracy on unseen trajectories. Qualitative results of the ground-truth tactile readings (collected by ZeMa) and the predicted tactile readings (output by Flow2Tactile Module) in simulation is shown in Figure~\ref{fig:zema_vis.pdf}.

\begin{figure}[!htbp]
    \centering
    \vspace{-0.5cm}
    \subfloat[]{
      \includegraphics[width=0.60225\linewidth]{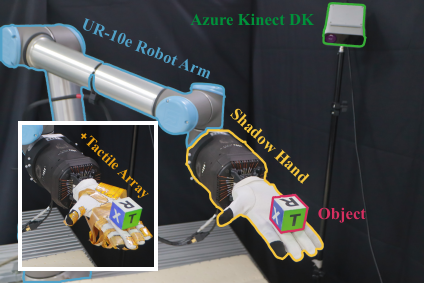}
      \label{fig:real_world_setting_left.pdf}
  }
    \subfloat[]{
      \includegraphics[width=0.39775\linewidth]{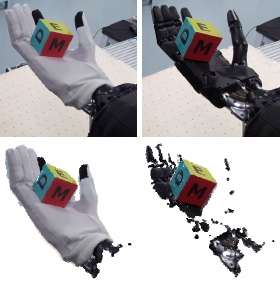}
      \label{fig:gloves.pdf}
  }
    \caption{(a): Real-world setup for both the vision-only mode and the visuotactile mode. (b): Images and point clouds were obtained by Kinect Azure before (right) and after (left) wearing gloves. The images are cropped versions that serve as input for image-based baselines. }
    \vspace{-0.3cm}
    \label{fig:real_world_setting.pdf}
\end{figure}
\subsection{Real-World Deployment} 
\subsubsection{Hardware Setup} 
\begin{figure}[!htbp]
    \vspace{-0.65cm}
    \centering
    \subfloat[]{
      \includegraphics[width=0.24\textwidth]{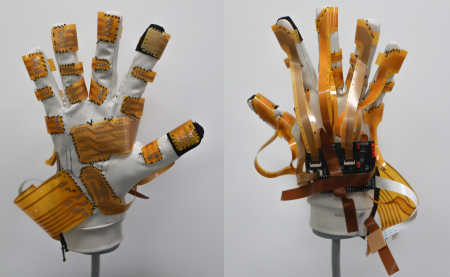}
      \label{fig:tactile_sensors.pdf}
  }
    \subfloat[]{
      \includegraphics[width=0.24\textwidth]{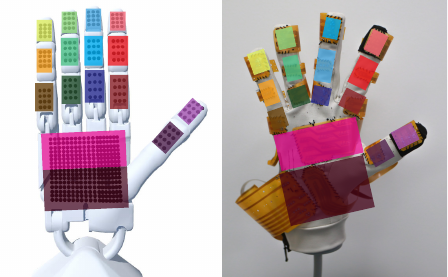}
      \label{fig:tactile_corresponse.pdf}
  }
    \caption{(a): The layout of the tactile sensors on the palm side (left) and the dorsal side (right) on the Shadow Hand in the real world. The tactile sensors are distributed on the palm side. (b): Regions for tactile readings mapping. Tactile sensors and contact keypoints are divided into 16 regions. Blocks of the same color indicate the corresponding regions.}
    \label{fig:tactile_sensor_and_regions}
    \vspace{-0.35cm}
\end{figure}

A Kinect Azure camera is used to capture visual observations. To mitigate metallic reflections degrading point clouds, we equip the Shadow Hand with a glove (Figure~\ref{fig:real_world_setting.pdf}). For the visuotactile mode, RunesKee tactile sensors (0$\sim$5N range, 0.1N resolution, 20 Hz communication frequency) are bought from \textit{Taobao}, molded on a flexible printed circuit (FPC), and attached to finger links and the palm. The layout of the sensor is customized to fit the robot hand, which is shown in Figure~\ref{fig:tactile_sensor_and_regions}(a). It features 8 sensors on the first phalanx of each finger, 3 on the second phalanx except for the thumb, and 6 on the third phalanx. In total, there are 82 sensors on the fingers, 66 sensors on the palm, bringing a total of 148 sensors. To fit our real sensor into the contact keypoints, we establish a mapping from the tactile sensors to the contact keypoints. Specifically, we divide the hand into several square regions, using the lower-left corner of each square region as the origin to calculate the coordinates of both the tactile sensors and contact keypoints. Then, for each contact keypoint, we select the tactile sensors within the same region and apply bidirectional linear interpolation to the force readings on these tactile sensors to finally obtain the tactile readings on the contact keypoints. The regions are shown in Figure~\ref{fig:tactile_sensor_and_regions}(b). 

\subsubsection{Sim-to-Real Transfer}
The FBI can be directly employed in real-world experiments. On the visual side, thanks to the glove coverage, the real-world point cloud quality is sufficient to take the direct transfer (Figure \ref{fig:real_world_setting.pdf}). On the tactile side, with a 
 correspondence between the real sensors and contact keypoints, we can easily convert sensor readings into contact keypoint readings, thanks to the binary tactile reading design. On the action side, to make the robot's behavior smoother, we apply an exponential moving average to the predicted actions and an action interpolation between the target joint pose and the current joint pose. After that, the FBI can successfully perform dexterous manipulation tasks in the real world. The target joint states are generated at about 20 Hz on average, accounting for both the time of point cloud transmission and the model inference time. 

\begin{figure*}[!htbp]
    \centering
    \vspace{-0.4cm}
    \includegraphics[width=1.0\linewidth]{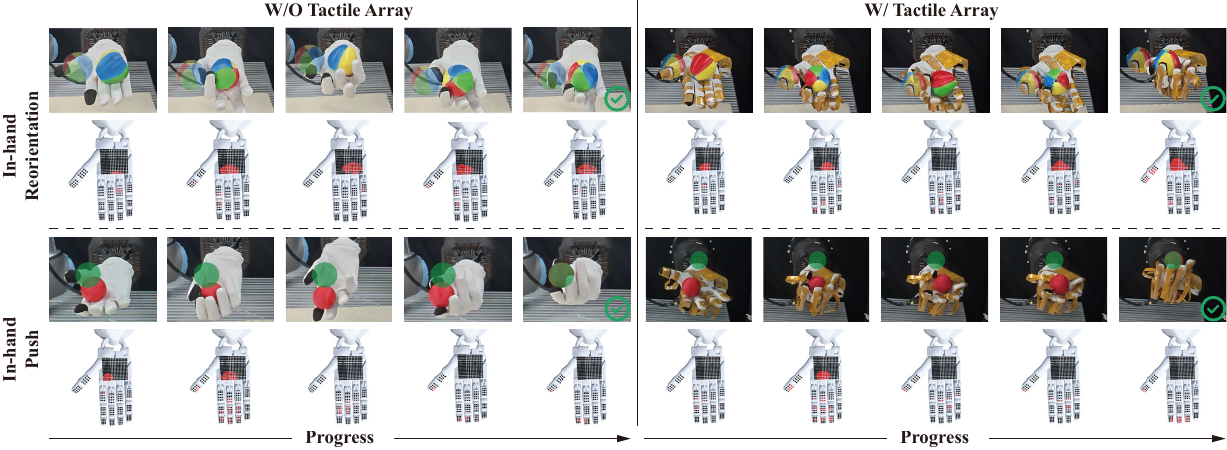}
    \caption{ \textbf{Qualitative results of the In-hand Reorientation task and the In-hand Push task.} We demonstrate both the vision-only mode (left) and the visuotactile mode (right). We capture and display five key frames from each manipulation process and the corresponding tactile readings. }
    \label{fig:qualitative result.png}
    \vspace{-0.5cm}
\end{figure*}
\subsection{Experiment Results}\label{sec:experiment results}

\subsubsection{Baselines} 
We compare against four baselines: DP3 (point cloud diffusion) \cite{ze20243d}, ManiCM (point cloud consistency model) \cite{lu2024manicm}, Ada-Flow (image-based flow matching) \cite{hu2024adaflow}, and Consistency Policy (image-based diffusion consistency) \cite{prasad2024consistency}, selected to emphasize visuotactile feature fusion benefits. For image-based methods, inputs are cropped to hand-object regions with uniform resolution.

\subsubsection{Simulation Results}
For simulation tasks, three seeds (0,1,2) are tested. Each seed evaluates 20 episodes every 20 epochs, with top-5 success rates averaged per seed. We report the mean and standard deviation of success rates across three seeds.  The simulation results presented in Table~\ref{table: simulation results} reveal that, when averaged over all tasks, FBI achieves 16.6\% (Vision-Only) to 18.4\% (Visuotactile) improvement compared to DP3, with a smaller variance. In the In-hand Reorientation task, FBI surpasses DP3 with a 19.3\% (Vision-Only) to 21.4\% (Visuotactile) improvement averaged over 9 different objects, indicating that after fusing dense contact states with visual information, our algorithm demonstrates superior performance, particularly in handling more complex dexterous tasks. With the tactile information, FBI can lower the chance of objects getting stuck during the process, reach the target more smoothly, and reduce the occurrence of objects slipping from the hand.

\subsubsection{Real-World Experiment Results} For real-world tasks, we evaluate each task with 20 trials and count the success rates. We keep hardware settings the same for every task to ensure fair comparison. Results in Table~\ref{table:real world experiments} indicate that our results in the simulation experiments remain similar in real-world settings. Moreover, it also shows that our algorithm, after sim-to-real processing, can be successfully applied to real-world scenarios. Similarly, with the implementation of tactile information and the shortcut model, the manipulation process in the real world becomes faster, smoother, and more reliable. Qualitative results shown in Figure~\ref{fig:qualitative result.png} illustrate two successful manipulation processes implemented in the real world. 

\subsection{Ablation Study}\label{sec:ablations}
In the ablation study, we analyze data scale impact, Flow2Tactile importance, tactile input variations, fusion method variations,  generalization, and shortcut model inference speed effects.

\subsubsection{Data Scale vs. Success Rate} To validate our data scale for contact-rich tasks, we analyze success rate vs. trajectory count using in-hand reorientation as a case study (Table~\ref{table:data-sr}). Considering the task complexity, it does not show promising results after 100 data samples. The influence of the data sample number converges after 5000 trajectories.

\begin{table}[!htbp]

\centering
\vspace{-0.04in}
\resizebox{\linewidth}{!}{%
\begin{tabular}{l|ccccccccc}
\toprule

    & \multicolumn{9}{c}{Number of Successful Trajectories}\\
   & 10& 100& 500& 1000& 3000& 4000& 5000& 6000 & 8000  \\
\midrule
\textbf{SR} & $0.0$& $4.3$& $16.0$& $27.6$& $45.1$& $51.8$& $55.6$& $56.2$& $56.6$  \\
\bottomrule
\end{tabular}}
\caption{Relationship between SR and the number of successful trajectories. We test FBI (Vision-Only) on the In-hand Reorientation task in simulation. SR is averaged across $9$ objects.}
\label{table:data-sr}
\vspace{-0.08in}
\end{table}

\subsubsection{Importance of Flow Prediction in Tactile Information Generation} 
We compare contact state generation methods to evaluate the tactile processing module: 1) Flow2Tactile (Ours): the contact states are generated by searching the pre-placed keypoints with the predicted flow; 2) Point Cloud (PC) to Tactile: the contact states are generated by searching the pre-placed keypoints with two-frame point clouds directly; 3) TOCH~\cite{zhou2022toch}: a hand-object contact modeling method based on spatio-temporal correspondences, yet suffers from heavy optimization and slow real-time inference 4) Ground Truth: We use the ground-truth tactile readings $R_{gt}$ (mentioned in Section~\ref{sec:f2t}) calculated in simulation. Results are shown in Table~\ref{table:Flow To Tactile}.

\begin{table}[!htbp]
\centering
\vspace{-0.02in}

\resizebox{\linewidth}{!}{%
\begin{tabular}{l|c|c|c|cc}
\toprule
    Algorithm $\backslash$ Task & \multicolumn{1}{c|}{\parbox{1.5cm}{\centering In-hand \\ Reorientation}} & \multicolumn{1}{c|}{\parbox{1.5cm}{\centering In-hand \\ Push}} & \multicolumn{1}{c|}{Adroit} & \textbf{Average} \\

\midrule

\textbf{Flow2Tactile (Ours)}  & \ccgf{55.6} & \ccgf{95.0} & 82.0 & \ccgf{64.7} \\

PC to Tactile  & $47.6$  & $90.0$ & $77.0$ & $57.6$ \\

TOCH & 53.9  & 94.0 & \ccgf{83.3} & 63.8 \\

\midrule

Ground Truth  & 58.3 & 97.0 & 84.0 & 67.2 \\

\bottomrule
\end{tabular}}
\caption{SR results on different tactile generation methods. We include the ground-truth tactile readings $R_{gt}$ as a reference. }
\label{table:Flow To Tactile}
\vspace{-0.15in}
\end{table}
\subsubsection{Different Forms of Tactile Information}
To demonstrate the benefits of the \textbf{dense contact} information, we conduct experiments on different forms of tactile information: 1) Dense Binary Contact (Ours): The contact states consist of binary tactile readings on 456 contact keypoints; 2) Dense Continuous Contact: The contact states consist of continuous tactile readings (force applied to the tactile sensors) on 456 contact keypoints; 3) Sparse Binary Contact: The contact states consist of link-wise binary tactile readings (readings between objects and the 24 links on the Shadow Hand); 4) W/o Contact: The policy model works without contact conditions. Results are shown in Table~\ref{table:different tactile taskwise}.
\begin{table}[!htbp]
\centering
\vspace{-0.02in}
\resizebox{\linewidth}{!}{%
\begin{tabular}{l|c|c|c|cc}
\toprule
    Algorithm $\backslash$ Task & \multicolumn{1}{c|}{\parbox{1.5cm}{\centering In-hand \\ Reorientation}} & \multicolumn{1}{c|}{\parbox{1.5cm}{\centering In-hand \\ Push}} & \multicolumn{1}{c|}{Adroit} & \textbf{Average} \\

\midrule

\textbf{Dense Binary (Ours)}  & \scgf{55.6} & \ccgf{95.0} & \scgf{82.0 }& \scgf{64.7}\\

Dense Continuous Contact  & \ccgf{56.7} & \scgf{93.0} & \ccgf{83.3} &  \ccgf{65.6}  \\

Sparse Contact  & $38.2$ & $87.0$ & $73.3$ & $50.1$ \\

W/o Contact  & $36.4$ & $87.0$ & $70.3$ & $48.1$   \\

\bottomrule
\end{tabular}}
\caption{Success Rate results on different tactile information. Lighter green indicates the second-best. }
\label{table:different tactile taskwise}
\vspace{-0.15in}
\end{table}

From Table~\ref{table:different tactile taskwise}, we can observe that dense tactile information, either binary or continuous, greatly enhances the model's performance in dexterous tasks. In contrast, joint-wise sparse tactile information only offers slight gains. Although dense continuous contact provides the best performance, we choose dense binary contact as our representation because it simplifies sim-to-real transfer and improves the accuracy of model predictions, as binary predictions are significantly easier for neural networks compared to regression.

\subsubsection{Benefits of the Transformer-based visuotactile fusing approach} To demonstrate the effectiveness of the visuotactile fusing module, we conduct experiments on different fusing methods: 1) Transformer Fusion (Ours): The visual and tactile features are forwarded to our transformer fusion module; 2) MLP Fusion: A multi-layer MLP is used to fuse the visuotactile features; 3) Add: We add the visual and tactile features to obtain the fused feature. Results are shown in Table~\ref{table:different fusion method}.
\begin{table}[!htbp]
\centering
\vspace{-0.2cm}
\resizebox{\linewidth}{!}{%
\begin{tabular}{l|c|c|c|cc}
\toprule
    Algorithm $\backslash$ Task & \multicolumn{1}{c|}{\parbox{1.5cm}{\centering In-hand \\ Reorientation}} & \multicolumn{1}{c|}{\parbox{1.5cm}{\centering In-hand \\ Push}} & \multicolumn{1}{c|}{Adroit} & \textbf{Average} \\

\midrule

\textbf{Transformer Fusion (Ours)} & \ccgf{55.6} & \ccgf{95.0} & \ccgf{82.0} & \ccgf{64.7} \\

MLP Fusion  & $48.1$ & $92.0$ & $77.3$ & $58.2$   \\

Add  & $43.2$ & $89.0$ & $75.7$ & $44.3$\\

\bottomrule
\end{tabular}}
\caption{Success Rate results on different fusing method Lighter green indicates the second-best. }
\label{table:different fusion method}
\vspace{-0.5cm}
\end{table}

\subsubsection{Generalization Ability to Unseen Objects} 
FBI demonstrates promising generalization to unseen object sizes/shapes. In cube reorientation, both FBI and DP3 succeed on original objects, but FBI outperforms DP3 on novel variants in zero-shot tests (Table~\ref{table:Generalization}), attributed to the contact modeling. We use FBI (Vision-Only) for fair comparison.

\begin{table}[!htbp]
\centering
\includegraphics[width=\linewidth]{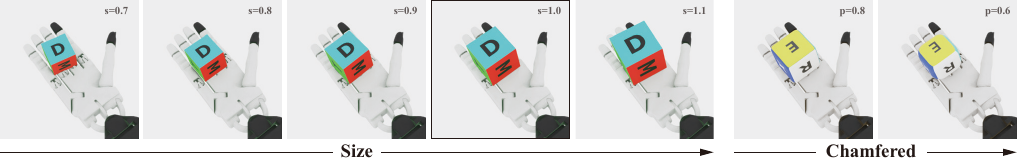}
\resizebox{1.0\linewidth}{!}{%
\begin{tabular}{l|l|ccccc}
\toprule

\parbox{2.4cm}{\centering Unseen Shape } & \parbox{1.2cm}{\centering Algorithms } & NO.1 & NO.2 & NO.3 & NO.4 & NO.5 \\

\midrule
     \parbox{2.4cm}{\centering $0.9 \times$ Size } & \parbox{1.2cm}{\centering \textbf{FBI (VO)} }  & \textcolor{ggreen}{\Checkmark} & \textcolor{ggreen}{\Checkmark} & \textcolor{ggreen}{\Checkmark} & \textcolor{ggreen}{\Checkmark} & \textcolor{ggreen}{\Checkmark} \\ 
 
 & \parbox{1.2cm}{\centering \textbf{DP3} } & \textcolor{ggreen}{\Checkmark} & \textcolor{ggreen}{\Checkmark} & \textcolor{ggreen}{\Checkmark} & \textcolor{gred}{\XSolidBrush} & \textcolor{ggreen}{\Checkmark} \\

\midrule
 
  \parbox{2.4cm}{\centering $0.8 \times$ Size } & \parbox{1.2cm}{\centering \textbf{FBI (VO)} } & \textcolor{ggreen}{\Checkmark} & \textcolor{ggreen}{\Checkmark} & \textcolor{gred}{\XSolidBrush} & \textcolor{ggreen}{\Checkmark} & \textcolor{ggreen}{\Checkmark} \\
  
  & \parbox{1.2cm}{\centering \textbf{DP3} } & \textcolor{gred}{\XSolidBrush} & \textcolor{ggreen}{\Checkmark} & \textcolor{gred}{\XSolidBrush} & \textcolor{gred}{\XSolidBrush} & \textcolor{ggreen}{\Checkmark} \\

\midrule
  
 \parbox{2.4cm}{\centering $0.75 \times$ Size } & \parbox{1.2cm}{\centering \textbf{FBI (VO)} } & \textcolor{ggreen}{\Checkmark} & \textcolor{gred}{\XSolidBrush} & \textcolor{gred}{\XSolidBrush} & \textcolor{gred}{\XSolidBrush} & \textcolor{ggreen}{\Checkmark}\\
 
 & \parbox{1.2cm}{\centering \textbf{DP3} }  & \textcolor{gred}{\XSolidBrush} & \textcolor{gred}{\XSolidBrush} & \textcolor{gred}{\XSolidBrush} & \textcolor{gred}{\XSolidBrush} & \textcolor{gred}{\XSolidBrush} \\

\midrule
 
 \parbox{2.4cm}{\centering $1.1 \times$ Size } & \parbox{1.2cm}{\centering \textbf{FBI (VO)} } & \textcolor{ggreen}{\Checkmark} & \textcolor{ggreen}{\Checkmark} & \textcolor{ggreen}{\Checkmark} & \textcolor{gred}{\XSolidBrush} & \textcolor{ggreen}{\Checkmark}\\
 
 & \parbox{1.2cm}{\centering \textbf{DP3} } & \textcolor{ggreen}{\Checkmark} & \textcolor{ggreen}{\Checkmark} & \textcolor{gred}{\XSolidBrush} & \textcolor{gred}{\XSolidBrush} & \textcolor{ggreen}{\Checkmark}\\

\midrule
 
 \parbox{2.4cm}{\centering Chamfered Cube \\ (Param.0.8)} & \parbox{1.2cm}{\centering \textbf{FBI (VO)} } & \textcolor{ggreen}{\Checkmark} & \textcolor{gred}{\XSolidBrush} & \textcolor{ggreen}{\Checkmark} & \textcolor{gred}{\XSolidBrush} & \textcolor{ggreen}{\Checkmark} \\
 
 & \hspace{0.5em} \textbf{DP3} & \textcolor{gred}{\XSolidBrush} & \textcolor{gred}{\XSolidBrush} & \textcolor{gred}{\XSolidBrush} & \textcolor{gred}{\XSolidBrush} & \textcolor{ggreen}{\Checkmark} \\

\midrule

  \parbox{2.4cm}{\centering Chamfered Cube\\(Param.0.6)}& \parbox{1.2cm}{\centering \textbf{FBI (VO)} } & \textcolor{ggreen}{\Checkmark} & \textcolor{gred}{\XSolidBrush} & \textcolor{gred}{\XSolidBrush} & \textcolor{gred}{\XSolidBrush} & \textcolor{gred}{\XSolidBrush}\\
 
 & \parbox{1.2cm}{\centering \textbf{DP3} } & \textcolor{gred}{\XSolidBrush} & \textcolor{gred}{\XSolidBrush} & \textcolor{gred}{\XSolidBrush} & \textcolor{gred}{\XSolidBrush} & \textcolor{gred}{\XSolidBrush} \\

\bottomrule
\end{tabular}}
\caption{\textbf{Size and shape generalization on cubes.} Each unseen size/shape is evaluated with 5 different initial and target poses, denoted No.1-No.5. VO stands for Vision-Only. }
\label{table:Generalization}
\vspace{-0.4cm}
\end{table}

 \begin{figure}[!htbp]
    \centering
    \vspace{-0.4cm}
    \includegraphics[width=1.0\linewidth]{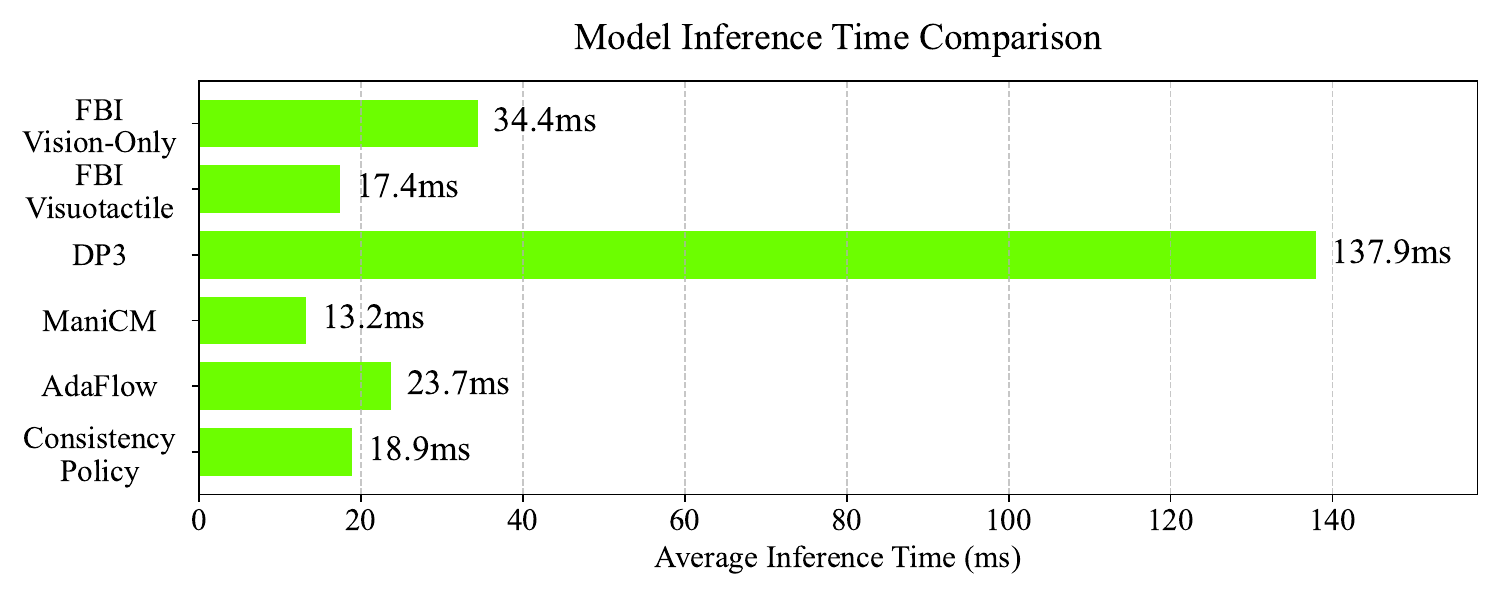}
    \caption{Average model inference time of all baselines in real-world experiments. We record the latency from receiving inputs (images/ point clouds/ robot states) to the final action series for all baselines for fairness. }
    \label{fig:inference_time_comparison.pdf}
    \vspace{-0.4cm}
\end{figure}

\subsubsection{Details on The Model's Inference Speed}
We now detail the inference speed in the real world and justify the use of shortcut models~\cite{frans2024one}. We report the \textbf{average model inference time} (denoted as AVG Time) for all baselines on an NVIDIA RTX 4090 GPU in real-world experiments in Figure~\ref{fig:inference_time_comparison.pdf}. FBI (Vision-Only) achieves 20 Hz control ($51.2\text{ms}$ per action series), including environment interaction ($16.8\text{ms}$) and model inference ($34.4\text{ms}$). it spends $17.4\text{ms}$ on the Flow2Tactile Module (flow prediction: $11.2\text{ms}$, tactile reading prediction: $6.2\text{ms}$) and $15.0\text{ms}$ on the visuotactile policy (Transformer fusion: $4.8 \text{ms}$, shortcut policy: $10.2\text{ms}$), with $2.0\text{ms}$ overhead. FBI (Visuotactile) reduces the model inference time to $17.4\text{ms}$ ($15.2\text{ms}$ for the policy and $2.2\text{ms}$ overhead) by skipping the Flow2Tactile Module. In contrast, DP3~\cite{ze20243d} requires $137.9\text{ms}$ for model inference, and even its simplified variant (Simple DP3) spends $83.6\text{ms}$. Replacing FBI’s shortcut models with 10-step DDIM~\cite{song2020denoising} would inflate latency to $261.4\text{ms}$, underscoring the necessity of our design.

\section{Conclusion}
\label{sec:conclusion}
This work introduces Flow Before Imitation (FBI), a visuotactile fusion framework that advances in-hand manipulation by dynamically linking tactile interactions to object motion dynamics. FBI’s dynamics-aware latent model enables synergy between vision and touch, supporting robust performance with or without physical tactile sensors. Evaluations across 5 dexterous tasks in both simulated and real-world settings demonstrate FBI’s superiority, achieving a 18.4\% higher success rate than state-of-the-art baselines, with notable gains (21.4\%) in in-hand reorientation. Critically, the FBI maintains stability under partial sensor failure, enhancing deployability in practical settings.









\bibliographystyle{IEEEtran}
\bibliography{IEEEabrv, reference}

\end{document}